\documentclass[runningheads]{llncs}
\usepackage{graphicx}
\usepackage{amsmath,amssymb} % define this before the line numbering.
\usepackage{color}
\usepackage{gensymb}
\usepackage{array}
\usepackage{amssymb}
\usepackage{wrapfig}
\usepackage{bm}
\usepackage{multirow}
\usepackage{multicol}
\usepackage{xspace}
\usepackage{diagbox}
\usepackage{caption}
\usepackage{subcaption}
\usepackage{float}
\usepackage{amsmath}
\usepackage{mathtools}

\begin{document}
\pagestyle{headings}
\mainmatter

\def\ACCV20SubNumber{253}  % Insert your submission number here

%===========================================================
\title{HDD-Net: Hybrid Detector Descriptor with Mutual Interactive Learning} % Replace with your title
\titlerunning{HDD-Net: Hybrid Detector Descriptor}
% If the paper title is too long for the running head, you can set
% an abbreviated paper title here
%
% \author{Axel Barroso-Laguna\inst{1}\orcidID{0000-0001-6933-3119} \and
% Yannick Verdie\inst{2}\orcidID{xxxx-xxxx-xxxx-xxxx} \and 
% Benjamin Busam\inst{2,3}\orcidID{xxxx-xxxx-xxxx-xxxx}
% \and
% Krystian Mikolajczyk\inst{1}\orcidID{0000-0003-0726-9187}}
% %
\author{Axel Barroso-Laguna\inst{1} \and
Yannick Verdie\inst{2} \and \\
Benjamin Busam\inst{2,3}  \and
Krystian Mikolajczyk\inst{1}}
\authorrunning{A. Barroso-Laguna et al.}
% First names are abbreviated in the running head.
% If there are more than two authors, 'et al.' is used.
%
\institute{Imperial College London \\
\email{\{axel.barroso17, k.mikolajczyk\}@imperial.ac.uk}
\and
Huawei Noah’s Ark Lab\\
\email{\{yannick.verdie, benjamin.busam\}@huawei.com}
\and
Technical University of Munich}

\maketitle
%%%%%%%%% BODY

%%%%%%%%% ABSTRACT
\begin{abstract}

Local feature extraction remains an active research area due to the advances in fields such as SLAM, 3D reconstructions, or AR applications. The success in these applications relies on the performance of the feature detector, descriptor, and its matching process. 
While the trend of detector-descriptor interaction of most methods is based on unifying the two into a single network, we propose an alternative approach that treats both components independently and focuses on their interaction during the learning process.
We formulate the classical hard-mining triplet loss as a new detector optimisation term to improve keypoint positions based on the descriptor map.  Moreover, we introduce a dense descriptor that uses a multi-scale approach within the architecture and a hybrid combination of hand-crafted and learnt features to obtain rotation and scale robustness by design. We evaluate our method extensively on several benchmarks and show improvements over the state of the art in terms of image matching and 3D reconstruction quality while keeping on par in camera localisation tasks.
\end{abstract}

%%%%%%%%% BODY TEXT
\section{Introduction}
\label{sec:Introduction}\noindent
At its core, a feature extraction method  identifies locations within a scene that are repeatable and distinctive, so that they can be detected with
high localisation accuracy under different camera conditions and be matched between different views. 
The results in vision applications such as image retrieval~\cite{teichmann2019detect}, 3D reconstruction~\cite{schonberger2016structure}, camera pose regression~\cite{sattler2019understanding}, or medical applications~\cite{busam2018markerless}, among others, have shown the  advantage of using sparse features over direct methods.

\begin{figure}[t]
 \centering
 \footnotesize
   \includegraphics[scale=0.32]{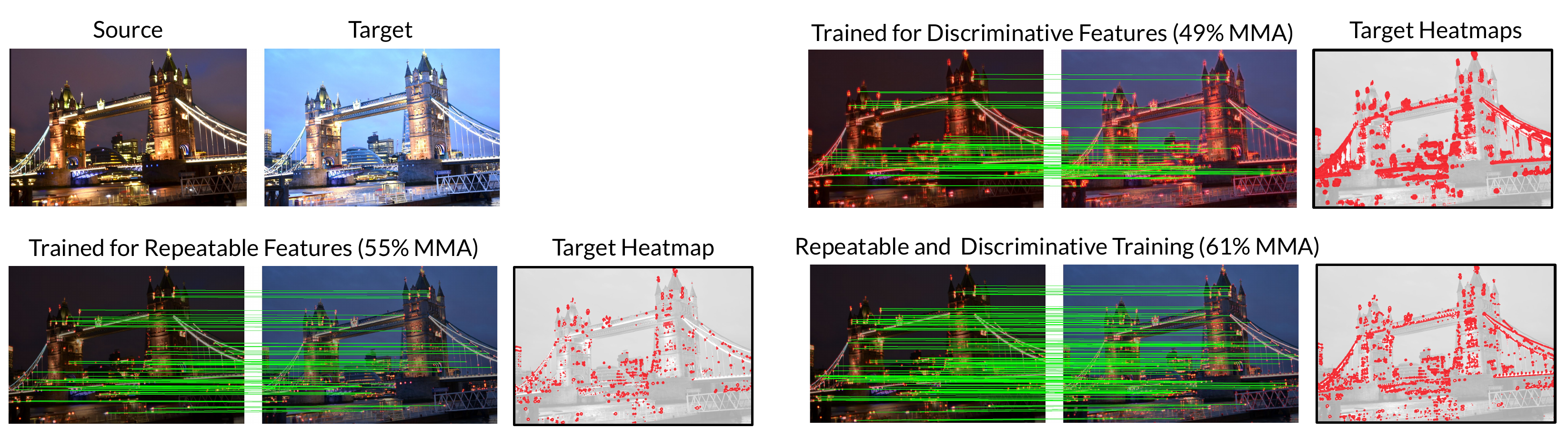}
    \caption{\textbf{Effect of different training strategies on the result.} Correct matches and target detection response maps on \textit{London Bridge} sequence (HPatches) when optimising the detector's features to be repetitive, discriminative, or both.}
    \label{fig:response_maps}
\end{figure}

Classical methods independently compute keypoints and descriptors. For instance, SIFT~\cite{lowe2004distinctive} focused on finding blobs on images and extracting gradient histograms as descriptors. Recently proposed descriptors, especially the patch-based  \cite{tian2017l2,mishchuk2017working}, are often trained for DoG keypoints~\cite{lowe2004distinctive}, 
and although they may perform well with other detectors~\cite{imwb2020}, their performance can be further improved if the models are trained with patches extracted by the same detector.
Similarly, detectors can benefit by training jointly with their associated descriptor~\cite{revaud2019r2d2}. Therefore, following the trend of using the descriptor information to infer the detections~\cite{dusmanu2019d2,revaud2019r2d2,yi2016lift}, we reformulate the descriptor hard-mining triplet cost function~\cite{mishchuk2017working} as a new detector loss.
The new detector term can be combined with any repeatability loss, and consequently, keypoint locations can be optimised based on the descriptor performance jointly with the detector repeatability. This approach leads to finding in a single score map both, repeatable and discriminative features, as shown in figure \ref{fig:response_maps}. We extend the network trainings to a multi-scale framework, such that the detector/descriptor learns to use different levels of detail when making predictions.\par

Our two-networks approach is motivated by the observations that jointly learnt detector-descriptor 
models~\cite{dusmanu2019d2,detone2018superpoint} lack keypoint localisation accuracy, which is critical for SLAM, SfM, or pose estimations \cite{imwb2020}, and the fact that keypoints are typically well localised on simple structures such as edges or corners, while descriptors require more context to be discriminative. We argue that despite the recent tendency for end-to-end and joint detector-descriptor methods, separate extractors allow for shallow models that can perform well in terms of accuracy and efficiency,  which has recently been observed in \cite{imwb2020}.
Besides that, in contrast to patch-based descriptors, dense image descriptors make it more difficult to locally rectify the image regions for invariance. To address this issue, we introduce an approach based on a block of hand-crafted features and a multi-scale representation within the descriptor architecture, making our network robust to small rotations and scale changes. We term our approach as HDD-Net: Hybrid Detector and Descriptor Network.\\

\noindent
In summary, our contributions are:
\begin{itemize}
  \item[$\bullet$] A new detector loss based on the hard-mining triplet cost function. Although the hard-mining triplet is widely used for descriptors, it has not been adapted to improve the keypoint detectors.
  \item[$\bullet$] A novel multi-scale sampling scheme to jointly train both architectures at multiple scales by combining local and global detections and descriptors.
  \item[$\bullet$] We improve the robustness to rotation and scale changes with a new dense descriptor architecture that leverages hand-crafted features together with multi-scale representations.
\end{itemize}

\section{Related Work}
\label{sec:Related_Work}\noindent
We focus the review of related work on learnt methods, and refer to~\cite{TuytelaarsMikolajczyk2007,csurka2018handcrafted,imwb2020,balntas2017hpatches,Karel_Vedaldi_BMVC_18,bojanic2019comparison} for further details. \\

\noindent
\textbf{Detectors.} Machine learning detectors were introduced with FAST~\cite{rosten2006machine}, a learnt algorithm to speed up the detection of corners in images.
Later, TILDE~\cite{verdie2015tilde} proposed to train multiple piecewise regressors that were robust under photometric changes in images.
DNET~\cite{lenc2016learning} and TCDET~\cite{zhang2017learning} based its learning on a formulation of the covariant constraint, enforcing the architecture to propose the same feature location in corresponding patches.
Key.Net~\cite{laguna2019key} expanded the covariant constraint to a multi-scale formulation, and used a hybrid architecture composed of hand-crafted and learnt feature blocks.
More details about the latest keypoint detectors can be found in \cite{Karel_Vedaldi_BMVC_18}, which provides an extensive detector evaluation. \\

\noindent
\textbf{Descriptors.} Descriptors have attracted more attention than detectors, particularly patch-based methods~\cite{balntas2016learning,tian2017l2,mishchuk2017working} due to the simplicity of the task and available benchmarks.
TFeat~\cite{balntas2016learning} moved from loss functions built upon pairs of examples to a triplet based loss to learn more robust representations.
In \cite{tian2017l2}, L2-Net architecture was introduced. L2-Net has been adopted in the following works due to its good optimisation and performance.
HardNet~\cite{mishchuk2017working} introduced the hard-mining strategy, selecting only the hardest examples as negatives in the triplet loss function.
SOSNet~\cite{tian2019sosnet} added a regularisation term to the triplet loss to include second-order similarity relationships among descriptors.
DOAP~\cite{he2018local} reformulated the training of descriptors as a ranking problem, by optimising the mean average precision instead of the distance between patches.
GeoDesc~\cite{luo2018geodesc} integrated geometry constraints to obtain better training data.\\
% Following the idea of improving the data, \cite{pultar2019leveraging} presented a new patch-based dataset containing scenes under different weather and seasonal conditions.\\
% Recently, HyNet~\cite{tian2020hynet} established a new state-of-the-art by combining different similarity functions and performing $L_2$ normalisation on the intermediate feature maps within the architecture. 

\noindent
\textbf{Joint Detectors and Descriptors.} 
LIFT~\cite{yi2016lift} was the first CNN based method to integrate detection, orientation estimation, and description.
LIFT was trained on quadruplet patches which were previously extracted with SIFT detector. 
SuperPoint~\cite{detone2018superpoint} used a single encoder and two decoders to perform dense feature detection and description. It was first pretrained to detect corners on a synthetic dataset and then improved by applying random homographies to the training images. 
This improves the stability of the ground truth positions under different viewpoints.
Similar to LIFT, LF-Net~\cite{ono2018lf} and RF-Net~\cite{shen2019rf} computed position, scale, orientation, and description.
LF-Net trained its detector score and scale estimator in full images without external keypoint supervision, and RF-Net extended LF-Net by exploiting the information provided by its receptive fields.
D2-Net~\cite{dusmanu2019d2} proposed to perform feature detection in the descriptor space,
showing that an already pre-trained network could be used for feature extraction even though it was optimized for a different task.
R2D2~\cite{revaud2019r2d2} introduced a dense version of the L2-Net~\cite{tian2017l2} architecture to predict descriptors and two keypoint score maps, which were each based on their repeatability and reliability.
ASLFeat~\cite{LuoASLFeat2020} proposed an accurate detector and invariant descriptor with multi-level connections and deformable convolutional networks~\cite{dai2017deformable,zhu2019deformable}.

\begin{figure}[t]
 \centering
   \includegraphics[scale=0.32]{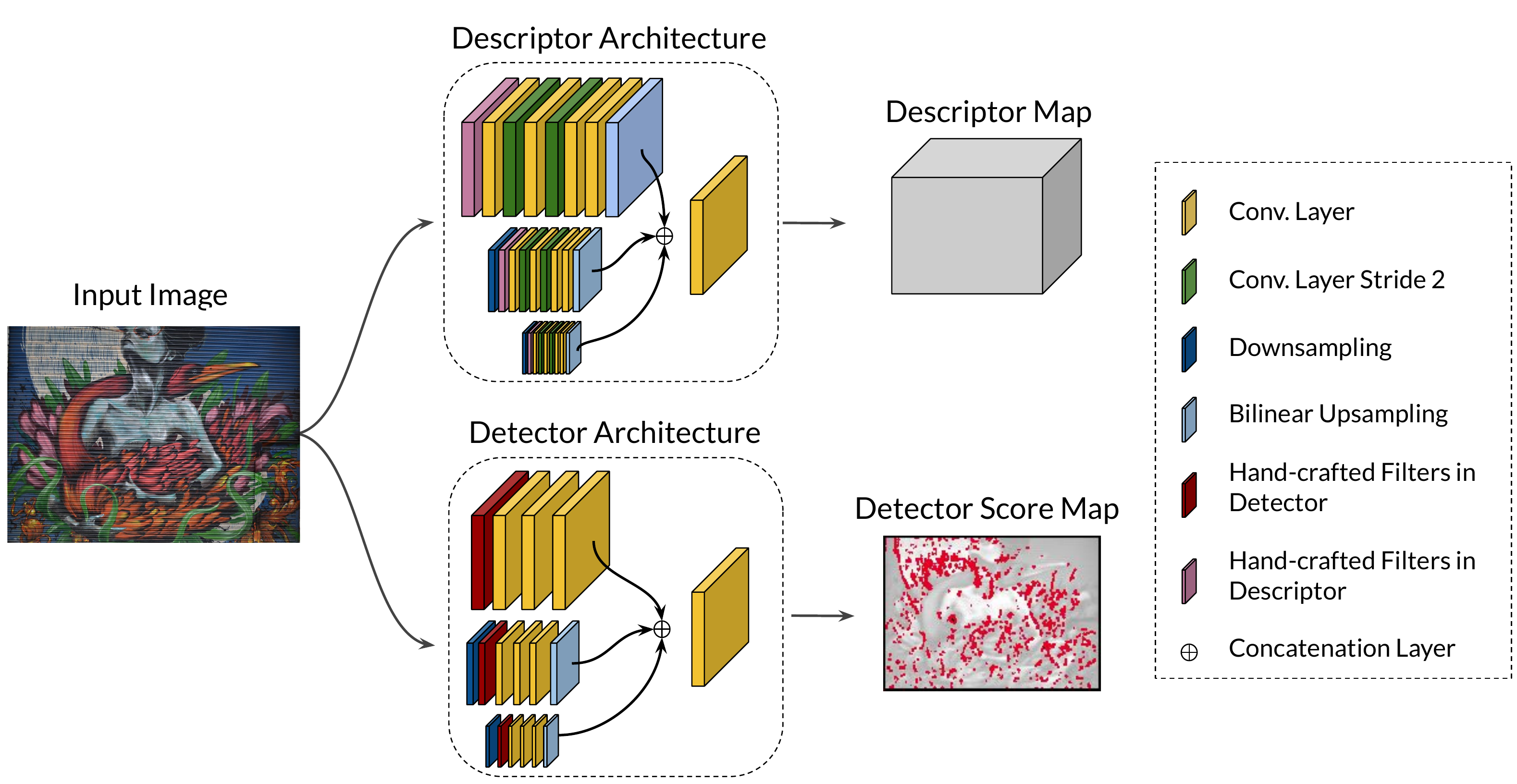}
    \caption{\textbf{HDD-Net Architecture}. HDD-Net is composed by two independent architectures. Instead of sharing a common feature extractor as in~\cite{detone2018superpoint,dusmanu2019d2,revaud2019r2d2,LuoASLFeat2020}, HDD-Net focuses its detector-descriptor interaction at the learning level.}
    \label{fig:full_method}
\end{figure}
\section{Method}
\label{sec:Method}
% This section presents the architecture and training of our HDD-Net.
\subsection{HDD-Net Architecture}
\label{sec:hddnet}
HDD-Net consists of two independent architectures for inferring the keypoint and descriptor maps, allowing to use different hand-crafted blocks that are designed specifically for each of these two tasks. 
Figure \ref{fig:full_method} shows the two independent blocks within the HDD-Net's feature extraction pipeline.\\

\noindent
\textbf{Descriptor.} 
As our method estimates dense descriptors in the entire image, an affine rectification of independent patches or rotation invariance by construction~\cite{ebel2019beyond} is not possible. To circumvent this, we design a hand-crafted block that explicitly addresses the robustness to rotation. We incorporate this block before the architecture based on L2-Net~\cite{tian2017l2}. 
As in the original L2-Net, we use stride convolutions to increase the size of its receptive field, however, we replace the last convolutional layer by a bilinear upsampling operator to upscale the map to its original image resolution. 
Moreover, we use a multi-scale image representation to extract features from resized images, which provides the network with details from different resolutions. After feature upsampling, multi-scale L2-Net features are concatenated and fused into a final descriptor map by a final convolutional layer. The top part of figure \ref{fig:full_method} shows the proposed descriptor architecture.\\

\noindent
\textbf{Rotation Robustness.} 
Transformation equivariance in CNNs has been extensively discussed in \cite{cohen2016group,follmann2018rotationally,worrall2019deep,dieleman2015rotation,marcos2017rotation}.
The two main approaches differ whether the transformations are applied to the input image~\cite{jaderberg2015spatial} or the filters~\cite{dieleman2016exploiting,marcos2017rotation}, we follow the latest methods and decide to rotate the filters. Rotating filters is more efficient since they are smaller than the input images, and therefore, have fewer memory requirements. 
\begin{figure}[t]
 \centering
   \includegraphics[scale=0.34]{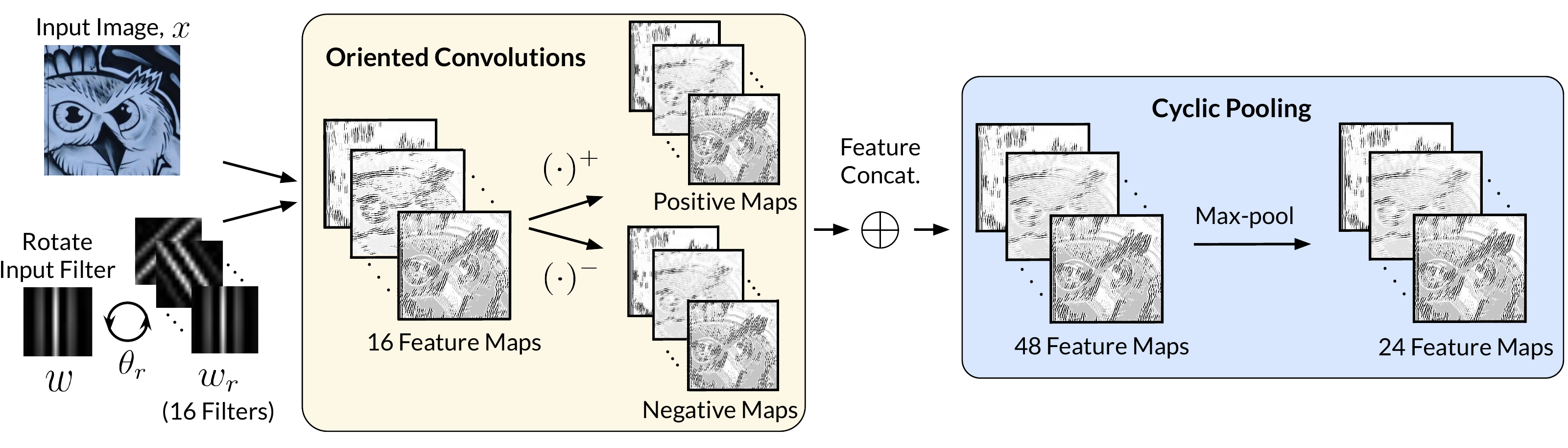}
    \caption{\textbf{Hand-crafted Block.} Given an input image, $x$, a designed filter, $w$, and a set of orientations, $\theta_r$, the rotation robustness is given by extracting and features from $x$ with each of the oriented filters, $w_r$. Additionally, $(\cdot)^+$ and $(\cdot)^-$ operators split positive and negative maps before the cyclic max-pooling block.}
    \label{fig:handcrafted_v3}
\end{figure}
Unlike \cite{dieleman2016exploiting}, our rotational filter is not learnt. We show in section \ref{subsec:Experiments_ablation} that the \mbox{pre-designed} filters offer a strong feature set that benefits the learning of consecutive CNN blocks.
Moreover, in contrast to \cite{dieleman2016exploiting}, which applies the rotation to all the layers in their convolutional model, we only focus on the input filter, which further reduces the computational complexity. However, we apply more rotations than \cite{dieleman2016exploiting} to the input filter to provide sufficient robustness. 
In \cite{marcos2017rotation}, authors proposed a method that applied multiple rotations to each convolutional filter.
Different than estimating a pixel-wise vector field to describe angle and orientation \cite{marcos2017rotation}, our rotation block returns multiple maxima through a cyclic pooling. 
The cyclic pooling operator returns local maxima every three neighbouring angles. We experimentally found that returning their local maxima provides better results than only using the global one. 
Thence, our hand-crafted block applies our input filter, $w$, at $R=16$ orientations, each corresponding to the following angles:
\begin{equation}
\theta_r =  \dfrac{360}{R} r \quad \textrm{and} \quad r \in [1, 2, ..., R].
\label{eq:rotations}
\end{equation}
A rotated filter is generated by rotating $\theta_r$ degrees around the input filter's center. Since our rotated filter is obtained by bilinear interpolation, we apply a circular mask to avoid possible artifacts on the filter's corners:
\begin{equation}
w_r =  m \cdot f(w, \theta_r),
\label{eq:rotations}
\end{equation}
with $m$ as a circular mask around filter's center and $f$ denoting the bilinear interpolation when rotating the filter, $w$, by $\theta_r$ degrees. Given an input image $I$, and our designed filter, $w$, we obtain a set of features $h(I)$ such as:
\begin{equation}
h_r(I) =  (I * w_r) \quad \textrm{and} \quad r \in [1, 2, ..., R],
\label{eq:rotations2}
\end{equation}
where $*$ denotes the convolution operator. Before the cyclic max-pooling block, and because max-pool is driven to positive values, we additionally split and concatenate the feature maps in a similar fashion to Descriptor Fields~\cite{crivellaro2014robust}:
\begin{equation}
\mathcal{H}_{r}(I) =  [ h_r(I), \textrm{ }(h_r(I))^{+}, \textrm{-1} \cdot (h_r(I))^{-}],
\label{eq:rotations3}
\end{equation}
with $(\cdot)^+$ and $(\cdot)^-$ operators respectively keeping the positive and negative parts of the feature map $h_r(I)$. Descriptor Fields proved to be effective under varying illumination conditions~\cite{crivellaro2014robust}. Our new set of features, $\mathcal{H}_{r}(I)$, are concatenated into a single feature map, $\mathcal{H}(I)$. Finally, we apply a cyclic max-pooling block on $\mathcal{H}(I)$. Instead of defining a spatial max-pooling, our cyclic pooling is applied in the channel depth, where each channel dimension represents one orientation, $\theta_r$, of the input filter. Cyclic max-pooling is applied every three neighbouring feature maps with a channel-wise stride of two, meaning that each feature map after max-pooling represents the local maxima among three neighbouring orientations. The full hand-crafted feature block is illustrated in figure~\ref{fig:handcrafted_v3}.\\

\begin{figure}[t]
 \centering
   \includegraphics[scale=0.32]{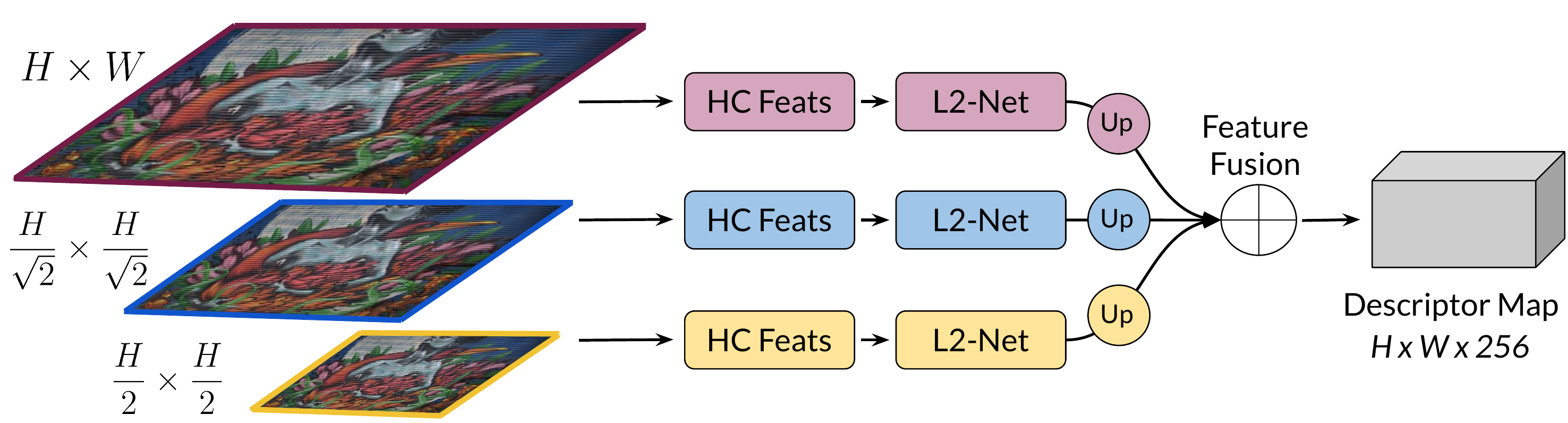}
    \caption{\textbf{Multi-Scale Hybrid Descriptor}. Gaussian pyramid is fed into our multi-scale descriptor. Each of the re-scaled input images go into one stream, which is composed by the hand-crafted block detailed in section \ref{sec:hddnet} and a L2-Net architecture. At the end, multi-scale L2-Net features are upsampled and combined through a final convolution.}
    \label{fig:descriptor_architecture}
\end{figure}

\noindent
\textbf{Scale Robustness.} Gaussian scale-space has been extensively exploited for local feature extraction~\cite{AKAZE,HarrisLaplace,yi2016lift}. In~\cite{ono2018lf,shen2019rf,laguna2019key}, the scale-space representation was used not only to extract multi-scale features but also to learn to combine their information. 
However, the fusion of multi-scale features is only used during the detection, while, in deep descriptors, it is either implemented via consecutive convolutional layers~\cite{detone2018superpoint} or by applying the networks on multiple resized images and combining the detections at the end~\cite{revaud2019r2d2,dusmanu2019d2,LuoASLFeat2020}. In contrast to~\cite{dusmanu2019d2,LuoASLFeat2020}, we extend the Gaussian pyramid to the descriptor part by designing a network that takes a Gaussian pyramid as input and fuses the multi-scale features before inferring the final descriptor. 
To fuse the extracted features, the network upsamples them into the original image resolution in each of the streams. Afterward, features are concatenated and fed into the last convolution, which maps the multi-scale features towards the desired descriptor size dimension as shown in figure~\ref{fig:descriptor_architecture}.
The descriptor encoder shares the weights on each multi-scale branch, hence, boosting its ability to extract features robust to scale changes. \\
% Figure~\ref{fig:descriptor_architecture} depicts the multi-scale descriptor.  \newline

\noindent
\textbf{Detector.} We adopt the architecture of Key.Net~\cite{laguna2019key} as shown in figure \ref{fig:full_method}. Key.Net combines specific hand-crafted filters for feature detection and a multi-scale shallow network. It has recently shown to achieve the state of the art results in repeatability \cite{laguna2019key,imwb2020}. Key.Net extended the covariant loss function proposed in \cite{lenc2016learning} to a multi-scale level, which was termed \mbox{M-SIP}. \mbox{M-SIP} splits the input images into smaller windows of size \mbox{$s_1 \times s_1$} and formulates the loss as the difference between soft-argmaximum positions in corresponding regions. 
\mbox{M-SIP} repeats the process multiple times but splitting the images each time with different window sizes, \mbox{$s_n \times s_n$}. The final loss function proposed by \mbox{M-SIP} between two images, $A$ and $B$, with their matrix transformation, $H_{b,a}$, is computed as the loss of all windows from all defined scale levels:
\begin{equation}
\mathcal{L}_{M-SIP}(A, B) = \sum_{\substack{i}} \| [{u_i}, {v_i}]^T_{a} - H_{b,a}[{u_i}, {v_i}]^T_{b}\|^2.
\label{eq:ms_msip_loss}
\end{equation}
We refer to \cite{laguna2019key} for further details.

\begin{figure}[t]
 \centering
   \includegraphics[scale=0.42]{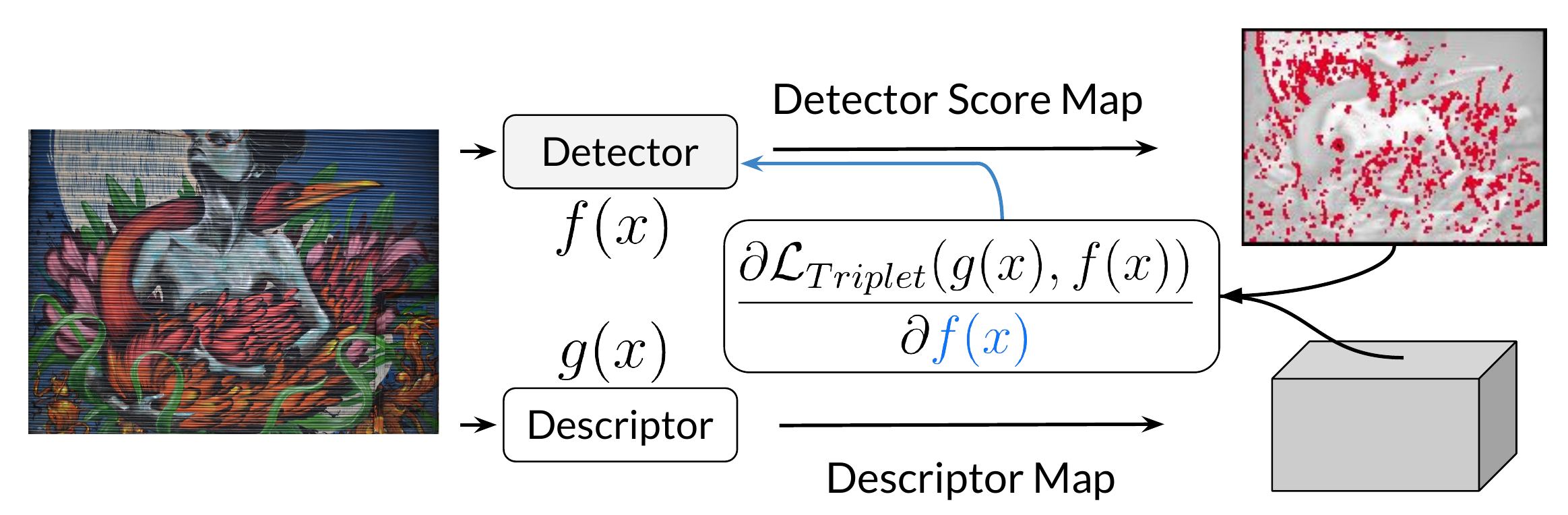}
\caption{\textbf{Detector-Descriptor Interaction.} The proposed triplet loss detector term optimises keypoint locations based on the descriptor map, refining the feature candidates towards more discriminative positions.}
\label{fig:triplet_learning_detector}
\end{figure}

\subsection{Descriptor-Detector Training}
\noindent
The detector learning has focused on localising features that are repeatable in a sequence of images~\cite{detone2018superpoint,ono2018lf,shen2019rf,verdie2015tilde,Karel_Vedaldi_BMVC_18,laguna2019key}, with a few works that determine whether these features are adequate for the matching stage~\cite{mishkin2018repeatability,revaud2019r2d2,yi2016lift,dusmanu2019d2}. 
Since a good feature should be repeatable as well as discriminative~\cite{TuytelaarsMikolajczyk2007}, we formulate the descriptor triplet loss function as a new detector learning term to refine the feature candidates towards more discriminative positions. 
Unlike AffNet~\cite{mishkin2018repeatability}, which estimates the affine shape of the features,  we refine only their locations, as these are the main parameters that are often used for end tasks such as SfM, SLAM, or AR. R2D2~\cite{revaud2019r2d2} inferred two independent response maps, seeking for discriminativeness of the features and their repeatability. Our approach combines both objectives into a single detection map. LIFT~\cite{yi2016lift} training was based on finding the locations with closest descriptors, in contrast, we propose a function based on a triplet loss with a hard-negative mining strategy. D2-Net~\cite{dusmanu2019d2} directly extracts detections from its dense descriptor map, meanwhile, we use Key.Net \cite{laguna2019key} architecture to compute a score map that represents repeatable as well as discriminative features.\\
% Note that even though some joint methods are trained end-to-end, and therefore, detector and descriptor learning affect each other, their keypoint detector does not directly address the discriminativeness of the features~\cite{detone2018superpoint,ono2018lf,shen2019rf}. \\

\begin{figure}[t!]
 \centering
   \includegraphics[scale=0.29]{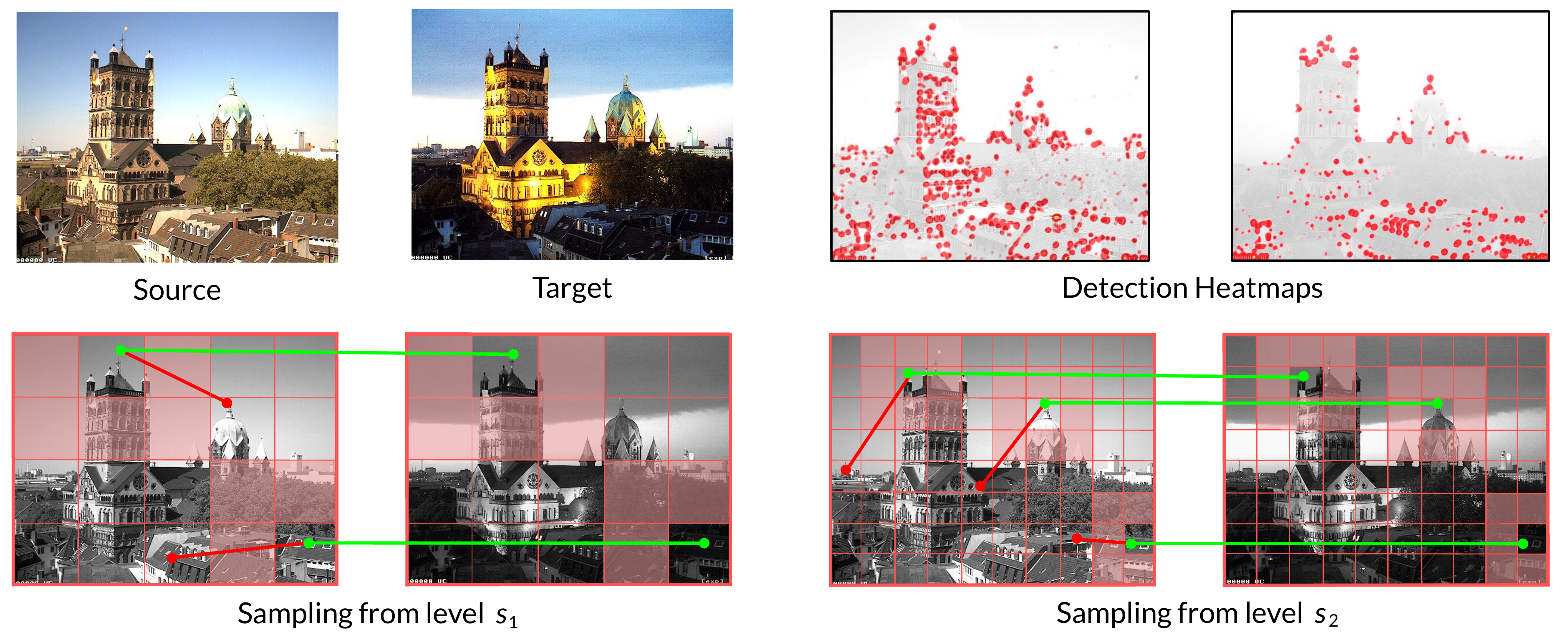}
    \caption{\textbf{Triplet Formation Pipeline.} Soft-descriptors are extracted from each window together with their respective positives and the hardest negatives. The negatives are extracted only from non-neighbouring areas (non-red areas).
    }
    \label{fig:triplet_Sampling}
\end{figure}

\noindent
\textbf{Detector Learning with Triplet Loss.} Hard-negative triplet learning maximises the Euclidean distance between a positive pair and their closest negative sample. 
In the original work~\cite{mishchuk2017working}, the optimisation happens in the descriptor part, however, we propose to freeze the descriptor such that the sampling locations proposed by the detector are updated to minimise the loss term as shown in figure~\ref{fig:triplet_learning_detector}. Then, given a pair of corresponding images, we create a grid on each image with a fixed window size of $s_1 \times s_1$.
From each window, we extract a \mbox{soft-descriptor} and its positive and negative samples as illustrated in figure~\ref{fig:triplet_Sampling}. To compute the \mbox{soft-descriptor}, we aggregate all the descriptors within the window based on the detection score map, so that the final soft-descriptor and the scores within a window are entangled. 
Note that if \mbox{Non-Maximum} Suppression (NMS) was used to select the maximum coordinates and its descriptor, we would only be able to back-propagate through the selected pixels and not the entire map.
Consider a window $w$ of size $s_1 \times s_1$ with the score value $r_i$ at each coordinate $[u,v]$ within the window. A softmax provides:
\begin{equation}
p(u,v) = \dfrac{e^{r(u, v)}}{\sum_{\substack{j, k}}^{s_1} e^{r(j, k)}}.
\label{eq:distribution}
\end{equation}
\noindent
The window $w$ has the associated descriptor vector $d_i$ at each coordinate $[u,v]$ within the window.
We compute the \mbox{soft-score}, $\Bar{r}$, and \mbox{soft-descriptor}, $\Bar{d}$, as:

\begin{equation}
\Bar{r} =\sum_{\substack{u, v}}^{s_1} r(u, v) \odot p(u, v) \quad \textrm{and} \quad \Bar{d} =\sum_{\substack{u, v}}^{s_1} d(u, v) \odot p(u, v).
\label{eq:soft_desc}
\end{equation}

\noindent
We use L2 normalisation after computing the \mbox{soft-descriptor}.
Similar to previous works \cite{mishkin2015mods,shen2019rf}, we sample the hardest negative candidate from a \mbox{non-neighbouring} region.
This geometric constraint is illustrated in figure~\ref{fig:triplet_Sampling}.
We can define our detector triplet loss with \mbox{soft-descriptors} in window $w$ as:
\begin{equation}
\mathcal{L}(w) = \mathcal{L}(\delta^{+}, \delta^{-}, \Bar{r}, \mu) =  \Bar{r}  \  \ \textit{max}(0,\mu + \delta^{+} - \delta^{-}),
\label{eq:window}
\end{equation}
where $\mu$ is a margin parameter, and $\delta^{+}$ and $\delta^{-}$ are the Euclidean distances between positive and negative \mbox{soft-descriptors} pairs. Moreover, we weight the contribution of each window by its \mbox{soft-score} to control the participation of meaningless windows \textit{e.g.}, flat areas.
The final loss is defined as the aggregation of losses on all $N_1$ windows of size $s_1 \times s_1$:
\begin{equation}
\mathcal{L}_{Trip}(s_1) =  \sum_{\substack{n}}^{N_1} \mathcal{L}(w_n) =  \sum_{\substack{n}}^{N_1} \mathcal{L}(\delta^{+}_n, \delta^{-}_n, \Bar{r}_n, \mu).
\label{eq:detector_triplet_loss}
\end{equation}

\noindent
\textbf{Multi-Scale Context Aggregation. } We extend equation~\ref{eq:detector_triplet_loss} to a multi-scale approach to learn  features that are discriminative across a range of scales. Multi-scale  learning was used in keypoint detection \cite{laguna2019key,ono2018lf,shen2019rf}, however, we extend these works by using the multi-scale sampling strategy not only on the detector but also on the descriptor training. Thus, we sample local soft-descriptors with varying window sizes, $s_j$ with $j \in [1, 2, ..., S]$, as shown in figure~\ref{fig:triplet_Sampling}, and combine their losses with control parameters $\lambda_{j}$ in a final term:
\begin{equation}
\mathcal{L}_{MS-Trip} =  \sum_{\substack{j}}\lambda_{j} \mathcal{L}_{Trip}(s_j),
\label{eq:ms_triplet_loss}
\end{equation}
\newline
\noindent
\textbf{Repeatable \& Discriminative.} The detector triplet loss optimises the model to find locations that can potentially be matched. As stated in \cite{TuytelaarsMikolajczyk2007}, discriminativeness is not sufficient to train a suitable detector. Therefore, we combine our discriminative loss and the repeatability term \mbox{M-SIP} proposed in~\cite{laguna2019key} with  control parameter $\beta$ to balance their contributions: 
\begin{equation}
\mathcal{L}_{R \textrm{ }\& D} =  \mathcal{L}_{M-SIP} + \beta\mathcal{L}_{MS-Trip},
\label{eq:ms_msip_triplet_loss}
\end{equation}
\newline
\noindent
\textbf{Entangled Detector-Descriptor Learning.} We frame our joint optimisation strategy as follows. The detector is optimised by equation~\ref{eq:ms_msip_triplet_loss}, meanwhile, the descriptor learning is based on the hard-mining triplet loss~\cite{mishchuk2017working}. For the descriptor learning, we use the same sampling approach as in figure~\ref{fig:triplet_Sampling}, however, instead of sampling soft-descriptors, we sample a point-wise descriptor per window. The location to sample the descriptor is provided by an NMS on the detector score map. Hence, the descriptor learning is conditioned by the detector score map sampling, meanwhile, our triplet detector loss term refines its candidate positions using the descriptor space. The interaction between parts tightly couples the two tasks and allows for mutual refinement. We alternate the detector and descriptor optimisation steps during training until a mutual convergence is reached. 

\subsection{Implementation Details}
\label{sec:implementation_evaluation_details}
\noindent
\textbf{Training Dataset.} We synthetically create pairs of images by cropping and applying random homography transformations to ImageNet images \cite{krizhevsky2012imagenet}. The image's dimensions after cropping are $192 \times 192$, and the random homography parameters are: rotation $[-30\degree, 30\degree]$, scale $[0.5, 2.0]$, and skew  $[-0.6, 0.6]$. However, illumination changes are harder to perform synthetically, and therefore, for tackling the illumination variations, we use the AMOS dataset \cite{pultar2019leveraging}, which contains outdoor webcam sequences of images taken from the same position at different times of the year. We experimentally observed that removing long-term or extreme variations \textit{i.e.}, winter-summer, helps the training of HDD-Net. Thus, we filter AMOS dataset such that we keep only images that are taken during summertime between sunrise and midnight. We generate a total of $12,000$ and $4,000$ images for training and validation, respectively.\\
\begin{table}[t!]
\begin{center}
\small
\footnotesize
\begin{tabular}{
@{\hskip1pt}>{\centering\arraybackslash}p{56px}
@{\hskip1pt}>{\centering\arraybackslash}p{32px}
@{\hskip3pt}>{\centering\arraybackslash}p{35px}
@{\hskip3pt}>{\centering\arraybackslash}p{40px}
@{\hskip3pt}>{\centering\arraybackslash}p{50px}
@{\hskip3pt}>{\centering\arraybackslash}p{38px}
@{\hskip3pt}>{\centering\arraybackslash}p{38px}
|@{\hskip3pt}>{\centering\arraybackslash}p{30px}
}
        \multicolumn{8}{c}{}\\
        Dense-L2Net &
        $1^{st}$ Order  &
        $2^{nd}$ Order &
        Gabor Filter &
        Fully Learnt &
        $(\cdot)^+$ \& $(\cdot)^-$ &
        Multi-Scale & MMA (\%)  \\
        \hline
        \checkmark & - & - & - & \checkmark & - & - & 41.8\\
        \checkmark & - & - & - & - & - & - & 42.0 \\
        \checkmark & \checkmark & - & - & - & - & - & 42.5 \\
        \checkmark & - & \checkmark & - & - & - & - & 43.1\\
        \checkmark & - & - & \checkmark & - & - & - & 43.3\\
        \checkmark & - & - & - & - & - &  \checkmark & 43.4 \\
        \checkmark & - & - & \checkmark & - & \checkmark & - & 43.6\\
        \checkmark & - & - & \checkmark & - & - &  \checkmark & 44.1 \\
        \checkmark & - & - & \checkmark & - & \checkmark &  \checkmark & \textbf{44.5} \\
        % \hline
    \end{tabular}
\caption{\textbf{Ablation Study.} Mean matching accuracy (MMA) on Heinly dataset \cite{heinly2012comparative} for different descriptor designs. Best results are obtained with Gabor filters in the hand-crafted block, $(\cdot)^+$ and $(\cdot)^-$ operators, and multi-scale feature fusion.}
\label{table:Ablation_architecture}
\end{center}
\end{table}

\noindent
\textbf{HDD-Net Training and Testing.} Although the detector triplet loss function is applied to the full image, we only use the top $K$ detections for training the descriptor. We select $K=20$ with a batch size of $8$. Thus, in every training batch, there is a total of $160$ triplets for training the descriptor. On the detector site, we use $j = [8, 16, 24, 32]$, $\lambda_{j} = [64, 16, 4, 1]$, and set $\beta = 0.4$. The hyper-parameter search was done on the validation set. We fix HDD-Net descriptor size to a 256 dimension since it is a good compromise between performance and computational time. Note that the latest joint detector-descriptor methods do not have a standard descriptor size, while \cite{revaud2019r2d2} is derived from 128-d L2-Net \cite{tian2017l2}, the works in \cite{detone2018superpoint,ono2018lf} use 256-d and \cite{dusmanu2019d2} is 512-d. During test time, we apply a $15\times15$ NMS to select candidate locations on the detector score map. HDD-Net is implemented in TensorFlow 1.15 and is available on GitHub\footnote{https://github.com/axelBarroso/HDD-Net}. 
% The training concludes within 48 hours on a single GTX 1080Ti.

\section{Experimental Evaluation}
\label{sec:Experiments}\noindent
This section presents the evaluation results of our method in several application scenarios. Due to the numerous possible combinations of independent detectors and patch-based descriptors, the comparison focuses against end-to-end and joint detector-descriptor state of the art approaches. 
\subsection{Architecture Design}
\label{subsec:Experiments_ablation}\noindent
\textbf{Dataset.} We use the Heinly dataset \cite{heinly2012comparative} to validate our architecture design choices. 
We focus on its homography set and use only the sequences that are not part of HPatches \cite{balntas2017hpatches}.  We compute the Mean Matching Accuracy (MMA) \cite{mikolajczyk2005performance} as the ratio of correctly matched features within a threshold of 5 pixels and the total number of detected features. \\

\noindent
\textbf{Ablation Study.} We evaluate a set of hand-crafted  filters for extracting features that are robust to rotation. Specifically, $1^{st}$ and $2^{nd}$ order derivatives as well as a Gabor filter. Besides, we further test a fully learnt approach without the hand-crafted filters. We also report results showing the impact of splitting the hand-crafted positive and negative features. Finally, our multi-scale approach is tested against a single-pass architecture without multi-scale feature fusion. \\

\noindent
\textbf{Results} in table~\ref{table:Ablation_architecture} show that the Gabor filter obtains better results than $1^{st}$ or $2^{nd}$ order derivatives. Gabor filters are especially effective for rotation since they are designed to detect patterns under specific orientations. Besides, results without constraining the rotational block to any specific filter are slightly lower than the baseline. The fully learnt model could be improved by adding more filters, but if we restrict the design to a single filter, hand-crafted filter with $(\cdot)^+$ and $(\cdot)^-$ operators give the best performance. Lastly, a notable boost over the baseline comes from our proposed multi-scale pyramid and feature fusion within the descriptor architecture.
\begin{figure*}[t]
    \centering
    \includegraphics[scale=0.30]{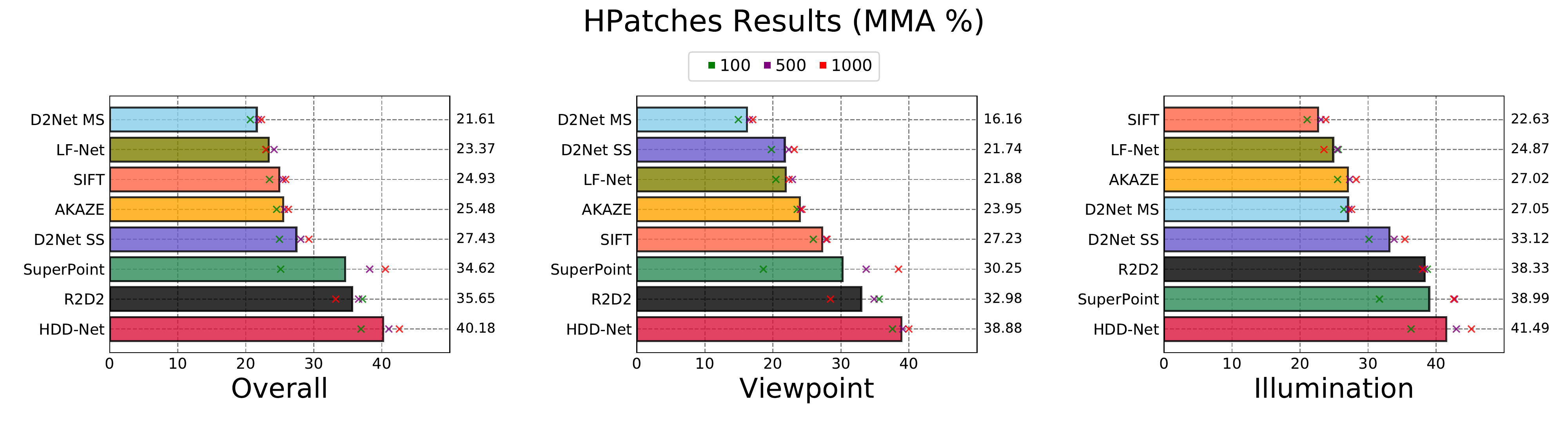}
    \caption{Mean Matching Accuracy (MMA) on HPatches dataset for top 100, 500 and 1,000 points. 
    HDD-Net gets the best results on both, viewpoint and illumination sequences.
    }
    \label{fig:matchingScore}
\end{figure*}
\begin{figure}[t]
    \centering
    \includegraphics[scale=0.40]{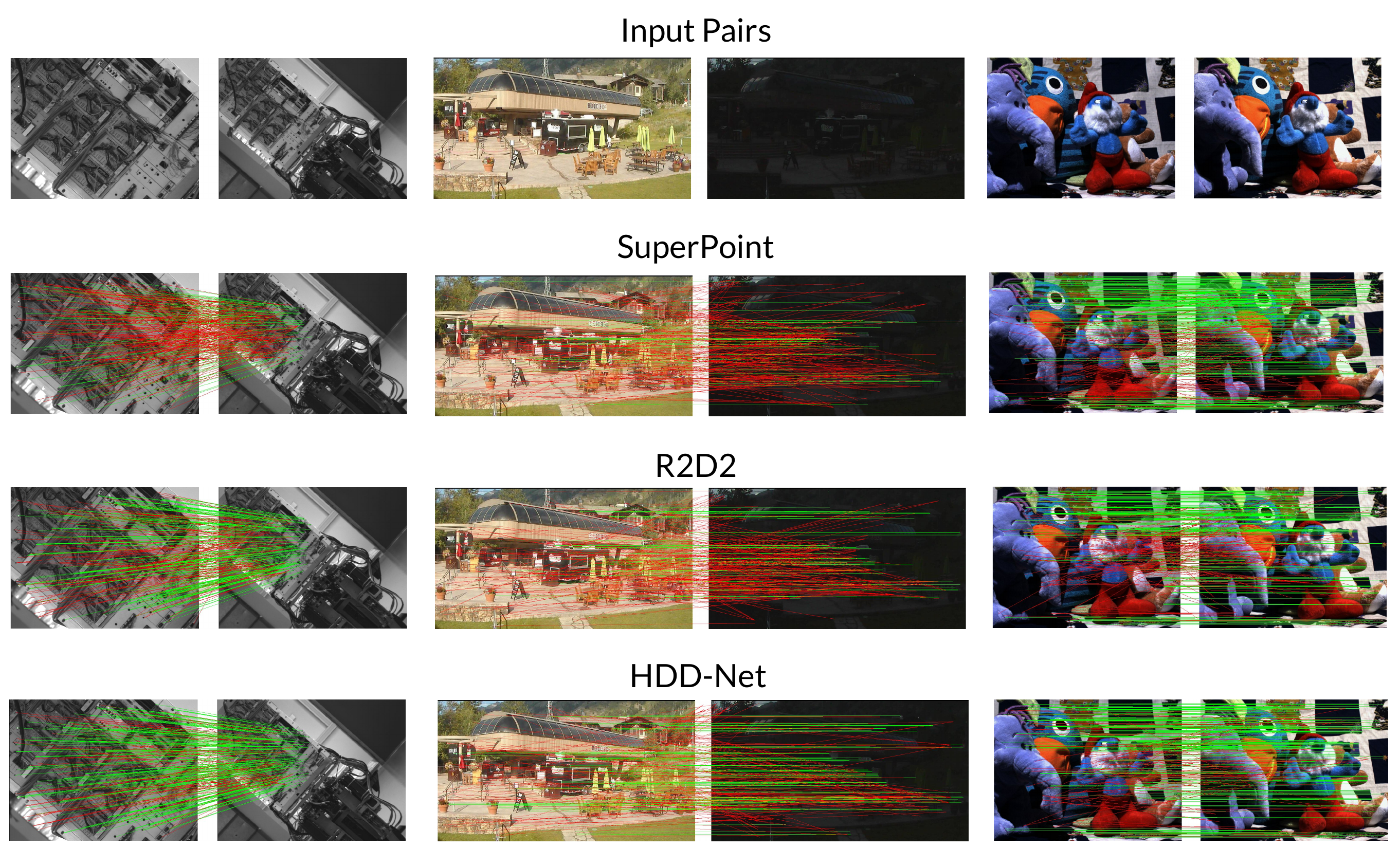}
    \caption{Qualitative examples on \textit{v\_bip}, \textit{i\_bridger}, and  \textit{i\_smurf} from the HPatches dataset. Illustrated sequences display extreme scale and rotation changes, as well as outdoor and indoor illumination variations.}
    \label{fig:qualitativeexamples}
\end{figure}
\subsection{Image Matching}
\noindent
\textbf{Dataset.} We use the HPatches~\cite{balntas2017hpatches} dataset with 116 sequences, including viewpoint and illumination changes. We compute results for sequences with image resolution smaller than \mbox{1200 $\times$ 1600} following the approach in \cite{dusmanu2019d2}. To demonstrate the impact of the detector and to make a fair comparison between different methods, we extend the detector evaluation protocol proposed in \cite{Karel_Vedaldi_BMVC_18} to the matching metrics by computing the MMA score for the top 100, 500, and 1,000 keypoints. As in section \ref{subsec:Experiments_ablation}, MMA is computed as the ratio of correctly matched features within a threshold of 5 pixels and the total number of detected features.\\
\begin{wraptable}{r}{0.40\textwidth}
\small
\begin{tabular}{lcc}
\noalign{\smallskip}
\multicolumn{1}{c}{} & \multicolumn{2}{c}{HPatches (MMA)} \\
\cline{2-3} \noalign{\smallskip}
\multicolumn{1}{c}{} & \multicolumn{1}{c}{View} & \multicolumn{1}{c}{Illum} \\
\noalign{\smallskip}
\hline
\noalign{\smallskip}
$\mathcal{L}_{MS-Trip}$ & 26.4 & 34.9 \\
$\mathcal{L}_{M-SIP}$ & 38.3 & 35.5 \\
$\mathcal{L}_{R \textrm{ }\& D} \textrm{ } (eq. \ref{eq:ms_msip_triplet_loss})$ & \textbf{38.9} & \textbf{41.5} \\
\end{tabular}
\caption{MMA (\%) results for different detector optimisations.
}
\label{table:Ablation_matchingScore}
\end{wraptable}
\noindent
\textbf{Effect of Triplet Learning on Detector.} Table~\ref{table:Ablation_matchingScore} shows HDD-Net results when training its detections to be repeatable ($\mathcal{L}_{M-SIP}$) or/and discriminative ($\mathcal{L}_{MS-Trip}$). The performance of $\mathcal{L}_{MS-Trip}$ only is lower than $\mathcal{L}_{M-SIP}$, which is in line with \cite{revaud2019r2d2}. Being able to detect repeatable features is crucial for matching images, however, best results are obtained with $\mathcal{L}_{R \textrm{ }\& D}$, which combines \mbox{$\mathcal{L}_{M-SIP}$ and $\mathcal{L}_{MS-Trip}$ with $\beta = 0.4$}, and shows the benefits of merging both principles into a single detection map.\\

\noindent
\textbf{Comparison to SOTA.} Figure~\ref{fig:matchingScore} compares our HDD-Net to different algorithms. HDD-Net outperforms all the other methods for viewpoint and illumination sequences on every threshold, excelling especially in the viewpoint change, that includes the scale and rotation transformations for which HDD-Net was designed. SuperPoint \cite{detone2018superpoint} performance is lower when using only the top 100 keypoints, and even though no method was trained with such constraint, the other models keep their performance very close to their 500 or 1,000 results. When constraining the number of keypoints, D2Net-SS \cite{dusmanu2019d2} results are higher than for its multi-scale version D2Net-MS.  D2Net-MS was reported in \cite{dusmanu2019d2} to achieve higher performance when using an unlimited number of features. In figure \ref{fig:qualitativeexamples}, we show matching results for the three best-performing methods on hard examples from HPatches. Even though those examples present extreme viewpoint or illumination changes, HDD-Net can match correctly most of its features.
% \begin{figure}[]
%     \centering
%     \includegraphics[scale=0.38]{gfx/sota_mma_th_2.pdf}
%     \caption{Qualitative examples on \textit{v_bip}, \textit{i_bridger}, and  \textit{i_smurf} from HPatches dataset. Illustrated sequences display extreme scale and rotation changes, as well as outdoor and indoor illumination variations.}
%     \label{fig:qualitativeexamples}
% \end{figure}
\begin{table}[t]
\begin{center}
\small
\begin{tabular}{
@{\hskip3pt}p{44px}
@{\hskip3pt}>{\centering\arraybackslash}p{20px}
@{\hskip3pt}>{\centering\arraybackslash}p{21px}
@{\hskip3pt}>{\centering\arraybackslash}p{21px}
@{\hskip3pt}>{\centering\arraybackslash}p{22px}
@{\hskip3pt}>{\centering\arraybackslash}p{1px}
@{\hskip3pt}>{\centering\arraybackslash}p{20px}
@{\hskip3pt}>{\centering\arraybackslash}p{21px}
@{\hskip3pt}>{\centering\arraybackslash}p{21px}
@{\hskip3pt}>{\centering\arraybackslash}p{22px}
@{\hskip3pt}>{\centering\arraybackslash}p{1px}
@{\hskip3pt}>{\centering\arraybackslash}p{20px}
@{\hskip3pt}>{\centering\arraybackslash}p{21px}
@{\hskip3pt}>{\centering\arraybackslash}p{21px}
@{\hskip3pt}>{\centering\arraybackslash}p{22px}
}
\noalign{\smallskip}
%  & \multicolumn{4}{c}{\textbf{Madrid Metropolis} (448 Images)} & & \multicolumn{4}{c}{\textbf{Gendarmenmarkt} (488 Images)} & & \multicolumn{4}{c}{\textbf{Tower of London} (526 Images)}\\ 
 & \multicolumn{4}{c}{\textbf{Madrid Metropolis}} & & \multicolumn{4}{c}{\textbf{Gendarmenmarkt}} & & \multicolumn{4}{c}{\textbf{Tower of London}}\\ 
  & \multicolumn{4}{c}{(448 Images)} & & \multicolumn{4}{c}{(488 Images)} & & \multicolumn{4}{c}{ (526 Images)}\\ 
%   & \multicolumn{4}{c}{(448 Images)} & & \multicolumn{4}{c}{(488 Images)} & & \multicolumn{4}{c}{526 Images}\\ 
\cline{2-5} \cline{7-10}  \cline{12-15} \noalign{\smallskip}
 & Reg. Ims & Sp. Pts & Track Len  & Rep. Err. & & Reg. Ims & Sp. Pts & Track Len  & Rep. Err.  & & Reg. Ims & Sp. Pts & Track Len  & Rep. Err. \\
 \cline{2-5} \cline{7-10} \cline{12-15} \noalign{\smallskip}
\noalign{\smallskip}
SIFT  & 27 & 1140 & 4.34 & 0.69  &  & 132 & 5332 & 3.68 & \textbf{0.86} &  & 75 & 4621 & 3.21 & 0.71\\
LF-Net & 19 & 467 & 4.22 & \textbf{0.62}  &  & 99 & 3460 & 4.65 & 0.90 &  & 76 & 3847 & 4.63 & \textbf{0.56}\\
SuperPoint & 39 & 1258 & 5.08 & 0.96  &  & \textbf{156} & \textbf{6470} & 5.93 & 1.21 &  & 111 & 5760 & 5.41 & 0.75\\
D2Net-SS & \textendash & \textendash & \textendash & \textendash  &  & 17 & 610 & 3.31 & 1.04 &  & 10 & 360 & 2.93 & 0.94\\
D2Net-MS & \textendash & \textendash & \textendash & \textendash  &  & 14 & 460 & 3.02 & 0.99 &  & 10 & 64 & 5.95 & 0.93\\
R2D2 & 22 & 984 & 4.85 & 0.88  &  & 115 & 3834 & \textbf{7.12} & 1.05  & & 81 & 3756 & \textbf{6.02} & 1.03\\
\hline
HDD-Net & \textbf{43} & \textbf{1374} & \textbf{5.25} & 0.80  & & 154 & 6174 & 6.30 & 0.98  & & \textbf{116} & \textbf{6039} & 5.45 & 0.80\\
\end{tabular}
\caption{3D Reconstruction results on the ETH 3D benchmark. Dash symbol ($ $\textendash $ $) means that COLMAP could not reconstruct any model. }
\label{tab:sfm_v2}
\end{center}
\end{table}
\subsection{3D Reconstruction}\noindent
\textbf{Dataset}. We use the ETH SfM benchmark \cite{schonberger2017comparative} for the 3D reconstruction task. We select three sequences; \textit{Madrid Metropolis}, \textit{Gendarmenmarkt}, and \textit{Tower of London}. We report results in terms of registered images (Reg. Ims), sparse points (Sp. Pts), track length (Track Len), and reprojection error (Rep. Err.). Top 2,048 points are used as in \cite{imwb2020}, which still provides a fair comparison between methods at a much lower cost.  The reconstruction is performed using COLMAP \cite{schonberger2016structure} software where we used one-third of the images to reduce the computational time. \\

\noindent
\textbf{Results}. Table~\ref{tab:sfm_v2} presents the results for the 3D reconstruction experiments. HDD-Net and SuperPoint obtain the best results overall. While HDD-Net recovers more sparse points and registers more images in \textit{Madrid} and \textit{London}, SuperPoint does it for \textit{Geendarmenmarkt}. D2-Net features did not allow to reconstruct any model on \textit{Madrid} within the evaluation protocol \textit{i.e.}, small regime on the number of extracted keypoints. Due to challenging examples with moving objects and in distant views, recovering a 3D model from a subset of keypoints makes the reconstruction task even harder. In terms of a track length,  
that is the number of images in which at least one feature was successfully tracked, 
 R2D2 and HDD-Net outperform all the other methods. \mbox{LF-Net} reports a smaller reprojection error followed by SIFT and HDD-Net. Although the reprojection error is small in \mbox{LF-Net}, their number of sparse points and registered images are below other competitors.

\subsection{Camera Localisation}\noindent

\noindent
\textbf{Dataset.} The Aachen Day-Night  \cite{sattler2018benchmarking} contains more than 5,000 images, with separate queries for day and night\footnote{We use the benchmark from the CVPR 2019 workshop on Long-term Visual Localization.}. Due to the challenging data, and to avoid convergence issues, we increase the number of keypoints to 8,000. Despite that, LF-Net features did not converge and are not  included in table~\ref{tab:camera_localization}. \\ 
\begin{wraptable}{r}{0.55\textwidth}
\small
\footnotesize
\begin{tabular}{lccc}
\noalign{\smallskip}
\multicolumn{1}{c}{} & \multicolumn{3}{c}{Aachen Day-Night} \\ 
\cline{2-4} \noalign{\smallskip}
\multicolumn{1}{c}{} & \multicolumn{3}{c}{Correct Localised Queries (\%)} \\ 
\cline{2-4} \noalign{\smallskip}
Threshold & 0.5m, $2^{\circ}$ & 1m, 5$^{\circ}$  & 5m, 10$^{\circ}$  \\
\noalign{\smallskip}
\hline
\noalign{\smallskip}
SIFT \cite{lowe2004distinctive} & 33.7 & 52.0 & 65.3 \\
SuperPoint \cite{detone2018superpoint} & 42.9 & 61.2 & 85.7 \\
D2-Net SS \cite{dusmanu2019d2} & 44.9 & 65.3 & \textbf{88.8} \\
D2-Net MS \cite{dusmanu2019d2} & 41.8 & \textbf{68.4} & \textbf{88.8} \\
R2D2 \cite{revaud2019r2d2} & \textbf{45.9} & 66.3 & \textbf{88.8} \\
\hline
HDD-Net & 43.9 & 62.2 & 82.7 \\
\end{tabular}
\caption{Aachen Day-Night results on localisation. The higher the better.}
\label{tab:camera_localization}
\end{wraptable}
\noindent\textbf{Results.} The best results for the most permissive error threshold are reported by D2-Net networks and R2D2. Note that D2-Net and R2D2 are trained on MegaDepth~\cite{li2018megadepth}, and Aachen datasets, respectively, which contains real 3D scenes under similar geometric conditions. In contrast, SuperPoint and HDD-Net use synthetic training data, and while they perform better on image matching or 3D reconstruction, their performance is lower on localisation. As a remark, results are much closer in the most restrictive error, showing that HDD-Net and SuperPoint are on par with their competitors for more accurate camera localisation. 
\section{Conclusion}
\label{sec:Conclusion}

In this paper, we have introduced a new detector-descriptor method based on a hand-crafted block and multi-scale image representation within the descriptor. Moreover, we have formulated the triplet loss function to not only learn the descriptor part but also to improve the accuracy of the keypoint locations proposed by the detector. We validate our contributions in the image matching task, where HDD-Net outperforms the baseline with a wide margin. Furthermore, we show through extensive experiments across different tasks that our approach outperforms or performs as well as the top joint detector-descriptor algorithms in terms of matching accuracy and 3D reconstruction, despite using only synthetic viewpoint changes and much fewer data samples for training.

 \noindent{\bf Acknowledgements.} This  research  was  supported  by  UK EPSRC IPALM project EP/S032398/1.
%%%%%%%%% Literature

\bibliographystyle{splncs}
\bibliography{egbib}

\end{document}